\documentclass[10pt,twocolumn,letterpaper]{article}

\usepackage{wacv}       

\usepackage{graphicx}
\usepackage{amsmath}
\usepackage{amssymb}
\usepackage{booktabs}
\usepackage{xcolor}
\usepackage{tabularx}
\usepackage{adjustbox}
\usepackage{caption}
\usepackage{subcaption}
\usepackage{multirow}
\usepackage{tikz}
\usepackage[pagebackref,breaklinks,colorlinks]{hyperref}

\usepackage{cleveref}

\crefname{section}{Sec.}{Secs.}
\Crefname{section}{Section}{Sections}
\Crefname{table}{Table}{Tables}
\crefname{table}{Tab.}{Tabs.}

\def\wacvPaperID{1428} 
\def\confName{WACV}
\def\confYear{2025}

\begin{document}

\title{
SegDesicNet: Lightweight Semantic Segmentation in Remote Sensing with Geo-Coordinate Embeddings for Domain Adaptation}


\newcommand{\authorskip}{\hspace{18mm}}
\author{
   \hspace{-3mm}Sachin Verma \hspace{13mm} Frank Lindseth  \hspace{13mm} Gabriel Kiss \\
{\tt\small \hspace{-1mm}sachin.verma@ntnu.no \hspace{12pt} frankl@ntnu.no  \hspace{9pt}  gabriel.kiss@ntnu.no}\\
\small{Norwegian University of Science and Technology (NTNU), Norway}\\
}
\maketitle
\begin{abstract}
Semantic segmentation is essential for analyzing high-definition remote sensing images (HRSIs) because it allows the precise classification of objects and regions at the pixel level. However, remote sensing data present challenges owing to geographical location, weather, and environmental variations, making it difficult for semantic segmentation models to generalize across diverse scenarios. Existing methods are often limited to specific data domains and require expert annotators and specialized equipment for semantic labeling. In this study, we propose a novel unsupervised domain adaptation technique for remote sensing semantic segmentation by utilizing geographical coordinates that are readily accessible in remote sensing setups as metadata in a dataset. To bridge the domain gap, we propose a novel approach that considers the combination of an image's location-encoding trait and the spherical nature of Earth's surface. Our proposed SegDesicNet module regresses the GRID positional encoding of the geocoordinates projected over the unit sphere to obtain the domain loss. Our experimental results demonstrate that the proposed SegDesicNet outperforms state-of-the-art domain adaptation methods in remote sensing image segmentation, achieving an improvement of approximately 6\% in the mean intersection over union (MIoU) with a $\sim$ ~27\% drop in parameter count on benchmarked subsets of the publicly available FLAIR \#1 dataset. We also benchmarked our method performance on the custom split of the ISPRS Potsdam dataset. Our algorithm seeks to reduce the modeling disparity between artificial neural networks and human comprehension of the physical world, making the technology more human-centric and scalable.
\end{abstract}

\section{Introduction}
\label{sec:intro}
Remote sensing technologies have gained widespread use and have significantly transformed the field by making high-resolution remote sensing images (HRSIs) more accessible for a broad range of applications. These HRSIs offer detailed information about the Earth's surface, capturing numerous spectral and spatial characteristics at unparalleled resolution levels, expanding the range of observations that can be made, and enabling the observation of a greater variety of objects. Precise land cover information derived from HRSIs is utilized in various planning, monitoring, and management applications across different sectors, for example, urban \cite{weng2012remote,wang2022mapping, shelestov2021extension} and environmental \cite{yu2018semantic}. The foundation for land cover and land use classification is rooted in semantic segmentation (SS) techniques \cite{wang2018understanding}. SS provides a comprehensive understanding of the spatial distribution of objects and their boundaries. This technique assigns class labels to every pixel in an image, specifically identifying the type of class represented by each pixel. Deep Neural Networks (DNNs) \cite{miikkulainen2024evolving} have exhibited outstanding performance in tasks concerning scene comprehension, including SS \cite{chen2017deeplab,wang2020deep,xie2021segformer,long2015fully}, given consistent data distribution across training and test sets. The difficulty in remote sensing lies within the SS task, given the substantial variability in scenes and styles present in remote sensing images. Variations in data distribution can result from factors such as diverse terrains, different weather conditions, various sensor techniques, and cultural and economic disparities\cite{xu2023universal,peng2022domain}. These discrepancies can critically impact the data distribution, and a model trained effectively on the original dataset may not perform optimally when faced with new data, even if the categories remain consistent. To address this issue, a commonly used approach is transfer learning, which involves leveraging the knowledge gained from the source domain to improve the performance in the target domain. By fine-tuning the pre-trained model on a smaller target domain dataset, the model can adapt to the specific characteristics and variations present in the target domain, thus mitigating the effects of the domain shift. However, this step can be challenging owing to the dynamic nature of the target domain. Semantic labeling of data, a crucial aspect of this process, is a resource-intensive operation that requires substantial investment in time, effort, and expertise. Moreover, maintaining an up-to-date and representative target domain dataset is an ongoing concern because the evolving nature of real-world data necessitates continuous adjustments to ensure the effectiveness of the model across changing scenarios. Given the abundance of remote sensing images that are available without labels, it is impractical to retrain models specifically for each target domain. 
Consequently, unsupervised domain adaptation (UDA) \cite{ganin2015unsupervised,kang2019contrastive,long2016unsupervised} has emerged as a solution for generating high-quality segmentation in the target domain without the need for any annotations, thereby offering a practical and efficient solution for adapting models to new domains in the field of remote sensing. The significance of UDA has been widely recognized and explored in the field of computer vision (CV). Generic CV models are frequently employed for remote sensing image analysis without fully considering the unique characteristics and peculiarities of Earth Observation (EO) data. Numerous attempts have been made to utilize this information, particularly the location context, in non-remote sensing setups \cite{tang2015improving, ayush2021geography, mac2019presence}.
In this article, we propose a novel and highly efficient convolutional neural network called SegDesicNet, drawing inspiration from previous works such as \cite{10208468, SALCEDOSANZ2020256, 8113128}. Our approach focuses on UDA for the SS of remote-sensing images.

SegDesicNet introduces a novel algorithm that utilizes geographical coordinates to align the source and target domains in an unsupervised manner. By integrating scene understanding derived from the cosine similarity of grid-based geographical coordinates projected over a unit sphere as the domain adaptation loss in our SegDesicNet model, we try to enhance the model's ability to learn and adapt to its surroundings effectively. our motivation is to achieve rational modeling of human scene understanding, which is an essential aspect of neural networks. The consistency of this algorithm, that is, employing geo-coordinates in a global reference system, increases its versatility and applicability across any region of the world in a remote sensing setup.
The contributions of this study are as follows: 
\begin{itemize}
\item In this study, we introduce an unsupervised domain adaptation technique for HRSI semantic segmentation, aiming to mimic human comprehension effectively. To achieve this, we integrated GRID positional encoding, inspired by \cite{abbott2014nobel}, and projected it onto a unit sphere to simulate the Earth's surface.
\item Our proposed approach outperforms state-of-the-art (SOTA) methods on the widely recognized FLAIR \#1 \cite{challenge} data subset \cite{10208468}, showing a notable mIoU improvement of over 6\%. Furthermore, we validated our approach using the ISPRS Potsdam \cite{potsdam} dataset and achieved superior results in the considered target domain. 
\item Notably, our network architecture exhibits a remarkable reduction in the parameter countcompared with existing state-of-the-art approaches. Specifically, our network has a 27\% lower parameter count, enabling more efficient and lightweight models without compromising performance. This efficiency is crucial
for practical applications because it reduces the computational requirements and allows for faster and more
resource-efficient inferences. 
\end{itemize}
\section{Related Work}
  \subsection{Semantic Segmentation}
The advancement of network architecture for SS has seen a notable transition from CNNs \cite{long2015fully,chen2017deeplab,zhao2017pyramid} to Vision Transformers \cite{chen2022vision,dosovitskiy2020image,xie2021segformer,yu2020context, zhao2017pyramid} over time. Another area of research has emerged that focuses on enhancing extracted representations by incorporating attention mechanisms \cite{huang2019ccnet, fu2019dual , li2018pyramid, zhong2020squeeze} or context representations \cite{yu2020context,wang2020deep, lin2018multi, yuan2021ocnet, yuan2020object, zhang2018context} into segmentation models. Within this context, our proposed approach, SegDesicNet, complements existing frameworks and consistently improves several state-of-the-art methods.
  \subsection{Unsupervised Domain Adaptation}
There are multiple approaches to achieving UDA\cite{peng2022domain}. The following methods can be considered, but they are not limited to: feature alignment\cite{abramov2020keep, sun2016deep, na2022contrastive, wang2021domain}, discriminate methods\cite{tsai2018learning,vu2019advent,vu2019dada,hoyer2022daformer, hoyer2022hrda} and labeling adjustment \cite{xie2020self,zou2018unsupervised,zhao2023learning}. The feature alignment approach focuses on aligning the features or characteristics of the source and target domains. The goal was to reduce the distribution discrepancy between the domains. Pseudo-labeling is utilized to adjust or refine the labels of the target domain. The model predictions on the target domain were treated as pseudo-labels, which were then used to enforce consistency in the predictions. Discriminative methods incorporate loss terms or mechanisms that encourage the model to distinguish between source and target domains. These methods aim to learn domain-invariant representations. Instance-based methods \cite{zhang2018importance} adapts instances or samples from the source domain to align them with the target domain. Hybrid approaches\cite{xie2023sepico} combine multiple strategies for a more comprehensive adaptation, integrating features from different domains, combining feature alignment, labeling adjustment, and discriminative methods. Self-supervised learning\cite{araslanov2021self} methods aims to reduce the domain gap between the source and target domains, enhance the adaptability of machine-learning models across different environments or datasets without the need for data labels, and is a widely used technique in computer vision for SS.
  \subsection{Unsupervised Domain Adaptation for Remote Sensing}
Li \etal \cite{rs14194942} showcased the effectiveness of the transformer \cite{vaswani2017attention} in self-training the UDA of remote sensing images, presenting two strategies, Gradual Class Weights and Local Dynamic Quality, to enhance the performance of the self-training UDA framework. Zhu \etal introduced MemoryAdaptNet\cite{zhu2023unsupervised}, which constructs an output space adversarial learning scheme and embeds an invariant feature memory module to store invariant domain-level context information, thereby improving the cross-domain semantic segmentation performance. Ma \etal \cite{8506613} aligned domains using centroid and covariance strategies, while Gross \etal \cite{gross2019nonlinear} addressed nonlinear effects and illumination variations by using labeled training spectra to align multiple hyperspectral datasets and navigate towards DNNs. Akiva \etal proposed MATTER\cite{Akiva_2022_CVPR} which learns material and texture representations. Manas \etal proposed SeCo\cite{Manas_2021_ICCV}, which utilizes an unsupervised acquisition procedure and self-supervised learning, whereas Ye \etal proposed UDAT\cite{Ye_2022_CVPR} which employs a unique object discovery approach to generate training patches. A transformer-based bridging layer was used to align the feature distributions in the night-time and daytime domains.

  \subsection{Geographical Metadata} 
First, geoinformation is used in the classification task\cite{Tang_2015_ICCV}, where it is concatenated into high-level image features of classification tasks to improve data distinctiveness. Subsequently, several similar methods were developed \cite{Salem_2020_CVPR,Workman_2021_ICCV,vivanco2023geoclip,Ayush_2021_ICCV}. Aodha \etal \cite{Aodha_2019_ICCV} modeled the presence of an object category as a function of geolocation and time. Inspired by neuroscience research, \cite{mai2020multi} introduced Space2Vec to encode the absolute positions and spatial relationships of places. \cite{russwurm2023geographic} lists a few possibilities for geographic location encoding. To our knowledge, only one specific work has addressed UDA with geo-coordinates for remote sensing\cite{10208468}, in which its DA module regresses the positional embedding of the geo-coordinates. The incorporation of weighted loss in training models involves updating the weights per image using an exponential moving average (EMA), which can lead to computationally intensive training. EMA in certain setups, such as ours, may not align naturally and could introduce unnecessary delays in convergence. This could result in overlooking or underemphasizing crucial updates, ultimately leading to a less accurate final model. The vector incorporates both latitude and longitude values, and the Mean Squared Error (MSE) may not be sufficient to account for the alignment or discrepancy in orientation and direction among the vectors. In contrast to the traditional Euclidean distance or MSE loss, cosine dissimilarity acknowledges that latitude and longitude values should be assessed on a spherical surface rather than in Euclidean space. The spherical nature of the Earth's surface is more accurately represented by cosine dissimilarity than other distance metrics that assume a flat Euclidean space. 

Building on this work, we extended the benchmark DA module of \cite{10208468} such that its learning ability is well utilized in capturing things the way they appear in the physical world, to make it more efficient for DA.

\begin{figure*}
  \centering
   \includegraphics[width=0.90\textwidth]{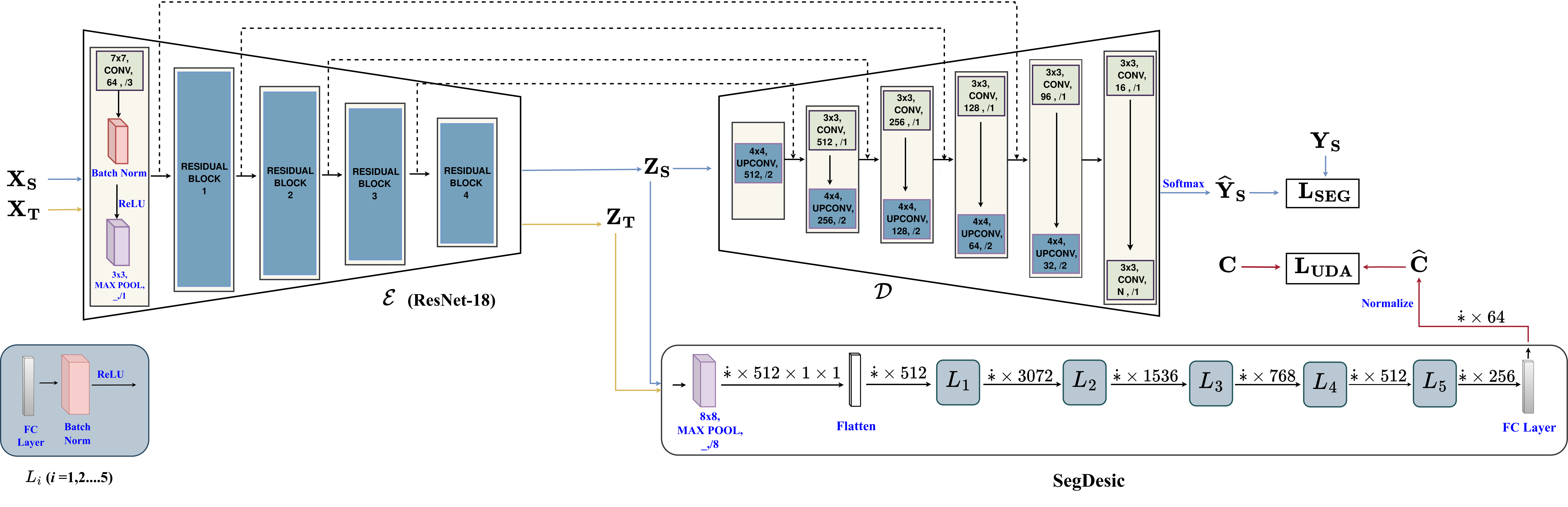}
  \caption{ Our model contains ResNet-18\cite{he2016deep} asencoder ($\mathcal{E}$), a decoder ($\mathcal{D}$) and the SegDesic module. The encoder and decoder together form a U-Net\cite{ronneberger2015unet}. The structure of the SegDesic module with its input and output sizes is shown on the lower block in the figure. In this diagram, $\dashrightarrow$ indicates its skip connections (with feature concatenation).\textit{CONV} represents a 2D convolution layer, \textit{UPCONV} signifies a 2D transposed convolution layer, \textit{$*\times *$} refers to the convolution kernel size, \textit{$*,/*'$} represents the number of kernels and convolution stride, \textit{$\dot*$} represents batch size, and $N$ represent number of class.
  }
  \label{fig:model}
\end{figure*}

\section{Methodology}
The datasets used in our experiment contain latitude and longitude information $(C_{lon}, C_{lat})$ for each image patch.
  \subsection{Model Architecture}
  \subsubsection{Baseline Semantic Segmentation Architecture}
Given a set of source images $\mathbf{X}_S \in \mathbb{R}^{H \times W \times B}$, where $H$ and $W$ are the height and width of the images, respectively, $B$ represents the number of color bands. Semantic labels accompany these images,
represented as $\mathbf{Y}_S$ with the same spatial dimension but a single band. We followed the lightweight U-Net\cite{ronneberger2015unet} as a baseline model to generate a semantic map from images using its RGB channel, similar to its ground truth. To accomplish this, we employed a pre-trained ResNet18\cite{he2016deep} model, which was originally trained on the Imagenet\cite{imagenet1} dataset, as the encoder in our U-Net implementation; we manipulated a few last layers in the decoder of this architectures, ending with an lightweight model with merely $\sim$ 24 million parameters. The last layer of the network utilizes softmax activation to produce the final output. This choice of encoder, coupled with softmax activation, allowed us to leverage the powerful feature extraction capabilities of ResNet18 while ensuring the generation of probabilistic predictions for the desired task. Our model is trained in a supervised manner using the renowned cross-entropy \cite{zhang2018generalized} loss. In our case, we termed this segmentation loss (\(L_{seg}\)), defined as 
\vspace{-1mm}
\begin{equation}
  L_{seg}=\frac{-\sum_{h=1}^{H} \sum_{w=1}^{W} y_{S}^{(h,w)} \cdot \log \left(h_{\theta}\left(x_{S}^{(h,w)}\right)\right)}{H\times W},
\end{equation}
where $h_{\theta}$ is the U-Net model with weights $\theta$ and $x_{S} \in \mathbf{X}_S$

  \subsubsection{SegDesic Module for Geographic Integration}

The SegDesic module depicted in \cref{fig:model}, is composed of six fully connected linear layers, each followed by batch normalization and ReLU activation, with the exception of the last layer. The last encoder layer of the U-Net architecture underwent max pooling to extract features, which were then passed as an input to the SegDesic module. The final output ($\widehat{\mathbf{C}}$) is a vector representing the geo-coordinates of the image patch, which is then normalized and 
cosine dissimilarity calculated between this final output and the normalized positional encoding ($\mathbf{C}$) of metadata $(C_{lon}, C_{lat})$ of the corresponding image is calculated, termed here as domain loss.


\subsection{Data Representation}
\subsubsection{Encoding Geographic Coordinates}

Considering the multi-scale periodic representation of grid cells in mammals and the established connection between spatial patterns and navigation, as explored by \cite{banino2018vector} and \cite{cueva2018emergence}, along with the demonstration of grid-like spatial patterns in trained networks, we propose a novel transformation to generate vectors that serve as supervision for $\widehat{\mathbf{C}}$. Our approach highlights the importance of grid cells for navigation and can be summarized like this: 
\begin{enumerate}
  \item Coordinate centering in the given EPSG:2154 
  reference system (for FLAIR \#1\cite{challenge} ). This ensures that the median values of the coordinates are $(0,0)$.
\vspace{-1.2mm}
    \begin{equation}
    \begin{aligned}
        C_{\text{lon}}' &= C_{\text{lon}} - 489353.59 \ \text{m} \\
        C_{\text{lat}}' &= C_{\text{lat}} - 6587552.20 \ \text{m}
    \end{aligned}
   \end{equation}

\item Transforming coordinates to the EPSG:4326 coordinate system, from EPSG:2154, a local coordinate system. This transformation aims to expand the algorithm's functionality and improve its ability to recognize geographical features in a standardized configuration. The transformation process is crucial for achieving optimal results in geographic analysis.
\begin{equation}
  \begin{aligned}
    \lambda , \phi = \text f_{2154\rightarrow4326}(C_{lon}',C_{lat}')
  \end{aligned}
  \end{equation}
\item Multi-scale approach adequately captures 2D positions, the positional encoding of coordinates is approached using a multi-scale way. This is necessary because a single scale representation alone is insufficient for accurately representing 2D positions due to the periodic nature of sine and cosine functions \cite{mai2020multi}.
\vspace{-1.4mm}
\begin{equation}
\label{eq:grid}
\small{
{\mathrm{GRID}}(\lambda,\phi)=\bigcup_{s=0}^{S-1}\left[\sin(\frac{\lambda}{\alpha_{s}}),\cos(\frac{\lambda}{\alpha_{s}}),\\sin(\frac{\phi}{\alpha_{s}}),\cos(\frac{\phi}{\alpha_{s}})\right] }
\end{equation} 

\vspace{-0.5 mm}where, $\alpha_{s} = \lambda_{m i n}\cdot g^{s/(S-1)}$ is a scaling factor that induces increasingly high frequencies through scales \(s\) from \(0\) to \(S-1\), \(\bigcup\) indicates concatenation and $\lambda_{m i n}$, $\lambda_{m a x}$ are the minimum and maximum grid scale with $g = \frac{\lambda_{m a x}}{\lambda_{m i n}}$.
 \item Projected this encoded vector onto a unit sphere. By mapping the encoded vector onto the unit sphere, we preserve the inherent spherical nature of Earth's observations (EO). When utilized as a ground truth, it allows the model to effectively capture and represent EO in their original spatial context.
 \vspace{-1.2mm}
 \begin{equation}
\label{eq:proj}
\mathbf{C} = \frac{{\mathrm{GRID}}(\lambda,\phi)}{\|{\mathrm{GRID}}(\lambda,\phi)\|_1}
\end{equation} 
\item Finally, we compute domain loss as 
 \vspace{-1.2mm}
 \begin{equation}
  L_{\text{UDA}} =1 - \frac{{\mathbf{C} \cdot \widehat{\mathbf{C}}}}{{\|\mathbf{C}\|_2 \cdot \|\widehat{\mathbf{C}}\|_2}} 
\end{equation}
\end{enumerate}

  \subsection{Domain Adaptation Strategy}
  \subsubsection{Domain Alignment Techniques}
A domain denoted by \(\mathbf{D}\) encompasses both the data (\(\mathbf{X}\)) and its corresponding distribution (\(\mathbf{P(x)}\)). Thus, the domain can be represented as \(\mathbf{D = \{X, P(x)\}}\), where \(\mathbf{x}\) is an element of \(\mathbf{X}\). Conversely, a task (\(\mathbf{T}\)) characterized by the label space (\(\mathbf{Y}\)) and prediction function (\(\mathbf{f(x)}\)) can be interpreted as the posterior probability \(\mathbf{p(y|x)}\). Hence, the task can be expressed as \(\mathbf{T = \{Y, P(y|x)\}}\), where \(\mathbf{y} \in\)  \(\mathbf{Y}\).\\
In domain adaptation (DA), it is typically assumed that the source domain (\(\mathbf{D_S}\)) and target domain (\(\mathbf{D_T}\)) share the same task (\(\mathbf{T_S = T_T}\)) but exhibit different distributions (\(\mathbf{P(x_s)} \neq \mathbf{P(x_t)}\)), where \(\mathbf{x_s} \in \) \(\mathbf{X_S}\) and \(\mathbf{x_t} \in\)  \(\mathbf{X_T}\). This indicates variations in the statistical properties between the data in the two domains, such as feature distributions or data generation processes. The objective of DA is to train a model that effectively transfers knowledge from the source domain to the target domain, despite these distributional differences, to enhance performance on the target domain task. Supervised and semi-supervised DA require labeled data in the target domain, whereas unsupervised DA operates under the assumption that there is no knowledge of the label space in the target domain (\(\mathbf{Y_T}\)).

  \subsubsection{Utilization of Geographic Features for Adaptation}
We adopted a similar architecture setup used by \cite{10208468} in our work, in which the input data $\mathbf{X}_S$ and $\mathbf{X}_T$ go into the encoder of U-Net processes denoted as $\mathcal{E}$ to produce feature maps $\mathbf{Z}_S \in \mathbb{R}^{H' \times W' \times C}$ and $\mathbf{Z}_T \in \mathbb{R}^{H' \times W' \times C}$, where $H'$, $W'$, and $C$ represent the height, width, and number of channels of the feature maps, respectively. Subsequently, decoder $\mathcal{D}$ of U-Net is fed with only $\mathbf{Z}_S$ to generate segmentation map $\hat{\mathbf{Y}}_S$. In addition, because the geo-coordinates are utilized for domain adaptation, both $\mathbf{Z}_S$ and $\mathbf{Z}_T$ are fed as inputs to the SegDesic module, which predicts a vector $\widehat{\mathbf{C}} \in \mathbb{R}^D$ containing the localization information of the image and is then normalized (in the same way as depicted in \cref{eq:proj}). The loss obtained from this module was used to fine-tune the encoder features. 
The final loss of SegDesicNet takes the following form:
\begin{equation}
\label{eq:fullloss}
L = L_{seg} + \alpha\times (L^{S}_{UDA} + L^{T}_{UDA})
\end{equation}
where ${L^S}_{UDA}$ and $L^{T}_{UDA}$ are the loss obtained from the SegDesic module for source and target data. To ensure that the contribution of the loss from different domains remains proportional to the segmentation loss, we performed scaling on its value using a hyperparameter \textit{$\alpha$}. 


\section{Experiment}






\vspace{-1.4 pt}
\begin{figure}[h!]
  \centering
  \begin{minipage}{0.50\linewidth}
    \centering
    
    \includegraphics[width = \linewidth]{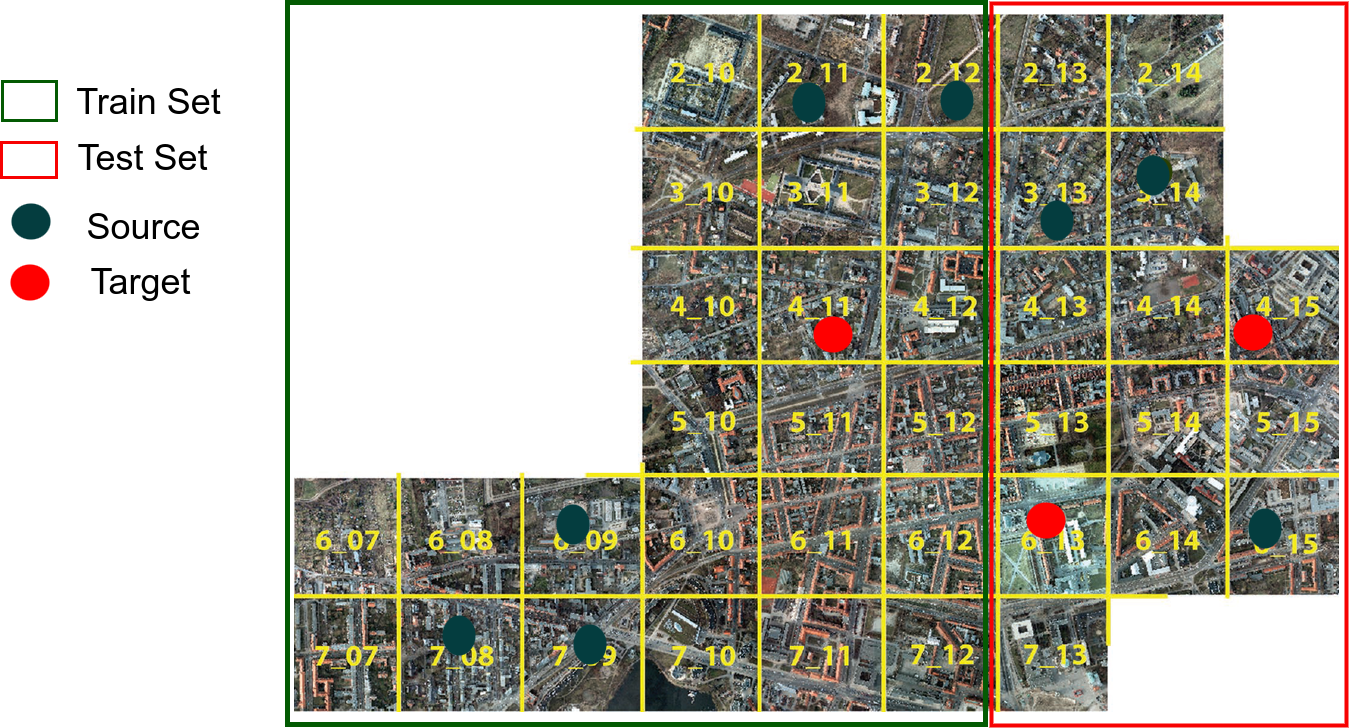}
  \end{minipage}\hfill
  \begin{minipage}{0.50\linewidth}
    \includegraphics[width=\linewidth ]{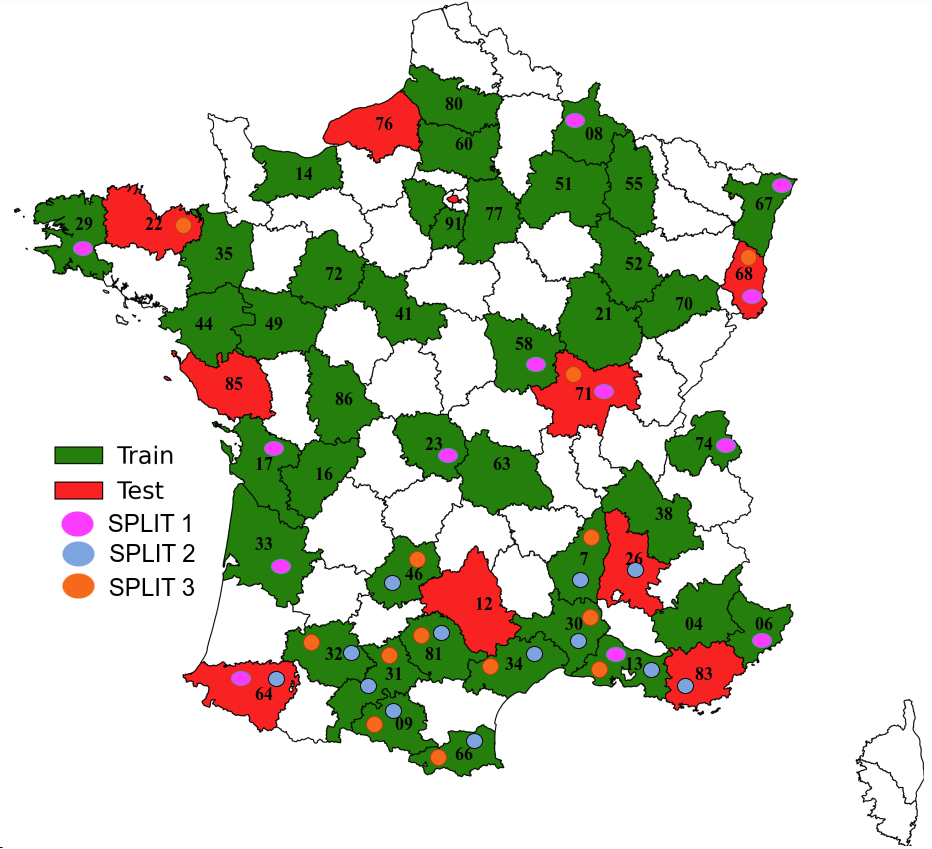}
  \end{minipage}
  \caption{Shown are the training and test set splits of the ISPRS Potsdam dataset (on the left) and the FLAIR \#1 dataset (on the right), with custom splits of the source and target domains indicated by colored dots. }
  \label{fig:dataset_split}
\end{figure}

\subsection{Datasets}
\textbf{FLAIR \#1}: It consists of 50 spatial domains, each depicting the diverse landscapes and climates of metropolitan France, corresponding to specific French departments \cite{challenge}. Each image patch measures 512 × 512 pixels with a ground sampling distance (GSD) of 0.2 meters, and each domain contains between 1,725 and 1,800 patches. Based on the implementation in \cite{10208468}, we used 12 major semantic classes and subsets for the source and target domain data in our experiments. In addition, we evaluated the performance of our model using different custom splits for the source and target domains. 
\\
\textbf{ISPRS Potsdam}: This dataset \cite{potsdam} consists of 38 labeled tiles, each covering an area of 6000 × 6000 pixels, with every pixel categorized into one of the six semantic classes. The dataset achieves a GSD of approximately 5 cm per pixel. Following standard practice mentioned in \cite{mmseg2024dataset}, the tiles were partitioned into non-overlapping images of 512 × 512 pixels.

The details of custom data splits are listed in \cref{table:splits} and the domain gap of in the data is presented in \cref{fig:dataset_split}. SPLIT1 is maintained as used in \cite{10208468}.

\begin{table*}[hbt!]
\begin{adjustbox}{width=\textwidth}
\footnotesize
\centering
\begin{tabular}{|c|c|p{7cm}|p{2cm}|c|c|}
\hline
\textbf{Data Set}& \textbf{SPLIT} & \textbf{Source Domain} & \textbf{Target Domain} & \textbf{\# Train Image} & \textbf{\# Test Image}\\ \hline
\multirow{3}{*}{FLAIR \#1}  &SPLIT1 \cite{10208468}&D06, D08, D13, D17, D23, D29, D33, D58, D67, D74 & D64, D68, D71& 16050 & 5350 \\
& SPLIT2 & D07, D09, D013, D30, D31, D32, D34, D46, D66, D81 & D26, D64, D83 &15325 & 5625\\ 
& SPLIT3 & D07, D09, D13, D30, D31, D32, D34, D46, D66, D81 & D22, D68, D71 &16272 &5300\\ 
\hline
ISPRS Potsdam & SPLIT4 &2\_11, 2\_12, 3\_13, 3\_14, 5\_11, 6\_15, 6\_9, 7\_8, 7\_9 & 4\_11, 6\_13, 4\_15 & 1296 & 432\\\hline

\end{tabular}
\end{adjustbox}
\caption{Custom Source and Target Domain Splits}
\label{table:splits}
\end{table*}

\begin{table*}[h!]
\centering

\begin{subtable}{\textwidth} 
\centering
\begin{adjustbox}{width=\textwidth}
\begin{tabular}{|lccccccccccccc|}
\hline
\textbf{Method} & \textbf{Building} & \textbf{Pervious} & \textbf{Impervious} & \textbf{Bare Soil} & \textbf{Water} & \textbf{Coniferous} & \textbf{Deciduous} & \textbf{Brushwood} & \textbf{Vine} & \textbf{Grassland} & \textbf{Crop} & \textbf{Plowed Land} & \textbf{mIoU} \\
\hline
CLAN\cite{clan} &6.24	&13.66	&17.09	&1.50	&12.99	&1.29	&27.22	&3.36	&30.69	&27.34	&7.69	&18.42&13.96 \\
AdaptSegNet \cite{Tsai_adaptseg_2018} & 39.98 & 20.75 & 40.23 & 20.36 & 15.25 & 4.93 & 35.37 & 10.99 & 34.51 & \textbf{42.69} & 11.06 & 23.47 & 24.97 \\
ADVENT \cite{vu2019advent} & 35.79 & 24.38 & 48.82 & 6.85 & 31.98 & 0.00 & 51.65 & 11.79 & 33.33 & 25.76 & 11.46 & 24.29 & 25.51 \\
IAST \cite{iast} &55.67	&36.43	&53.71	&26.95	&53.33	&0.00	&50.67	&11.56	&43.24	&26.28	&26.31	&44.27 &35.70 \\
DAFormer\cite{vu2019advent} & 67.09 & 45.56 & 61.99 & 55.35 & 65.12 & 8.91 & 54.39 & \textbf{20.31} & 64.39 & 38.79 & 23.74 & 41.83 & 45.62 \\
UDA\_for\_RS \cite{li2022unsupervised} & 66.30 & \textbf{48.05} & 62.36 & \textbf{59.28} & 61.24 & 9.22 & 60.02 & 16.52 & 57.74 & 40.12 & 30.32 & 54.17 & 47.11 \\
GeoMultiTaskNet \cite{10208468} & 67.53 & 40.86 & 63.89 & 55.31 & 67.02 & \textbf{13.85} & 60.97 & 14.08 & 53.09 & 40.33 & 35.02 & \textbf{54.79} & 47.23 \\
Ours & \textbf{67.65}&45.18 & \textbf{64.13} & 55.92 & \textbf{71.37} & 11.62 & \textbf{62.65} & 16.65 & \textbf{74.50} & 40.90 & \textbf{37.83} & 53.06 & \textbf{50.12} \\
\hline

\end{tabular}
\end{adjustbox}
\caption{SPLIT1}
\label{table:table1}
\end{subtable}

\hspace{5pt}

\begin{subtable}{\textwidth} 
\centering
\begin{adjustbox}{width=\textwidth}
\begin{tabular}{|p{3.7cm}ccccccccccccc|}
\hline
\textbf{Method} & \textbf{Building} & \textbf{Pervious} & \textbf{Impervious} & \textbf{Bare Soil} & \textbf{Water} & \textbf{Coniferous} & \textbf{Deciduous} & \textbf{Brushwood} & \textbf{Vine} & \textbf{Grassland} & \textbf{Crop} & \textbf{Plowed Land} & \textbf{mIoU} \\

\hline
CLAN\cite{clan} & 24.27 & 21.27 & 41.19 & 8.89 & 26.57 & 4.22 & 28.26 & 24.70 & 54.84 & 31.00 & 18.80 & 1.28 & 23.77 \\ 
AdaptSegNet \cite{Tsai_adaptseg_2018}& 60.07 & 24.06 & 54.22 & 21.29 & 24.31 & 11.90 & 39.02 & 26.31 & 47.25 & 31.56 & 19.95 & 10.56 & 30.88 \\ 
ADVENT\cite{vu2019advent} & 56.21 & 29.82 & 55.47 & 27.52 & 44.73 & 15.83 & 41.38 & 22.77 & 55.18 & 41.47 & 12.52 & 30.26 & 36.10 \\ 
IAST \cite{iast} & 68.84 & \textbf{42.05} & 61.19 & \textbf{61.86} & \textbf{79.38} & 0.00 & 46.27 & \textbf{35.69} & 69.40 & 35.38 & 34.06 & \textbf{36.71} & 47.57 \\ 
ours & \textbf{69.60} & 38.63 & \textbf{65.31} & 46.16 & 73.54 & \textbf{25.52} &\textbf{ 46.89} & 35.19 & \textbf{74.00} & \textbf{44.87} & \textbf{34.6}5 & 26.67 & 48.42 \\ 
\hline
\end{tabular}
\end{adjustbox}
\caption{SPLIT2}
\label{table:table2}
\end{subtable}

\hspace{5pt}

\begin{subtable}{\textwidth} 
\centering
\begin{adjustbox}{width=\textwidth}
\begin{tabular}{|p{3.7cm}ccccccccccccc|}
\hline
\textbf{Method} & \textbf{Building} & \textbf{Pervious} & \textbf{Impervious} & \textbf{Bare Soil} & \textbf{Water} & \textbf{Coniferous} & \textbf{Deciduous} & \textbf{Brushwood} & \textbf{Vine} & \textbf{Grassland} & \textbf{Crop} & \textbf{Plowed Land} & \textbf{mIoU} \\
\hline
CLAN\cite{clan} & 28.86 & 18.38 & 37.78 & 5.31 & 17.92 & 7.78 & 45.82 & 10.69 & 66.38 & 40.61 & 14.67 & 7.81 & 25.17 \\
ADVENT\cite{vu2019advent} & 49.15 & 30.39 & 50.13 & 18.23 & 43.70 & 15.75 & 59.95 & \textbf{15.37} & 49.89 & \textbf{53.45} & 9.18 & 22.87 & 34.84 \\
AdaptSegNet\cite{Tsai_adaptseg_2018} & 57.95 & 27.68 & 50.17 & 11.78 & 35.25 & 17.96 & 63.73 & 14.28 & 55.64 & 40.86 & 28.26 & 17.81 & 35.11 \\
IAST\cite{iast} & 62.31 & \textbf{52.13} & \textbf{64.67} & 15.04 & \textbf{75.28} & 0.00 & 61.19 & 12.70 & 69.00 & 30.85 & \textbf{39.85} & 49.28 & 44.36 \\
ours & \textbf{63.33} & 46.61 & 60.32 & \textbf{18.52} & 51.09 & \textbf{31.25} & \textbf{66.32} & 14.12 & \textbf{74.58} & 47.30 & 34.25 & \textbf{52.94} & \textbf{46.72} \\ 
\hline

\hline
\end{tabular}
\end{adjustbox}
\caption{ SPLIT3}
\label{table:table3}
\end{subtable}

\caption{Class-wise comparison of IoU(\%) for for different methods across various data domains of FLAIR \#1 dataset.}
\label{table:comparison}
\end{table*}

\begin{table*}[h!] 
 \centering
 \hspace{-0.2cm}
 \begin{minipage}{0.34\textwidth}
  \centering
  \begin{adjustbox}{height=1.35cm}
   \begin{tabular}{cc}
    \toprule
    \textbf{Architecture} & \textbf{\#Parameters(M)} \\
    \midrule
     AdaptSegNet \cite{Tsai_adaptseg_2018} &99\\
  ADVENT\cite{vu2019advent}& 99\\
  DAFormer\cite{Hoyer_2022_CVPR}&85\\
  UDA\_for\_RS \cite{li2022unsupervised}& 85 \\
  CLAN\cite{clan}& 44\\
  IAST\cite{iast} & 45 \\
  GeoMultiTaskNet \cite{10208468} & 33\\
  Ours & \hspace{-12pt}$\sim$ 24\\
    \bottomrule
   \end{tabular}
  \end{adjustbox}
  \caption{Comparing the parameter count of the baseline models used in SPLIT1 to that of ours}
  \label{table:model_param}
 \end{minipage}%
 \hspace{0.8cm}
 \begin{minipage}{0.6\textwidth}

   \centering 

  \begin{adjustbox}{height=1.34cm, width=10cm}

   \begin{tabular}{|ccccccc|}
    \hline
     \scriptsize {\textbf{Method}} &  \scriptsize{\textbf{Impervious}} &  \scriptsize{\textbf{Building} }& \scriptsize{\textbf{Low Vege}} & \scriptsize{\textbf{Tree}} & \scriptsize{\textbf{Car} }&  \scriptsize{\textbf{mIoU}} \\
  \hline
 \scriptsize{IAST\cite{iast}} & \scriptsize{68.42 }& \scriptsize{75.86} & \scriptsize{64.77} & \scriptsize{64.46} & \scriptsize{43.72} & \scriptsize{63.46 }\\
 \scriptsize{ CLAN\cite{clan}} & \scriptsize{64.67} & \scriptsize{68.71} & \scriptsize{58.72} & \scriptsize{47.28} & \scriptsize{42.63} & \scriptsize{56.40} \\
  \scriptsize{ADVENT\cite{vu2019advent} }& \scriptsize{72.18} & \scriptsize{78.33} & \scriptsize{69.65} & \scriptsize{67.65} & \scriptsize{68.69} & \scriptsize{71.30} \\
  \scriptsize{AdaptSegnet\cite{Tsai_adaptseg_2018}} & \scriptsize{73.40} & \scriptsize{78.89} & \scriptsize{68.14} & \scriptsize{67.69} & \scriptsize{70.65} & \scriptsize{71.75} \\
  
  \scriptsize{Ours} & \scriptsize{\textbf{74.88}} & \scriptsize{\textbf{81.17}} & \scriptsize{\textbf{70.49}} & \scriptsize{\textbf{69.47}} & \scriptsize{\textbf{72.26}} & \scriptsize{\textbf{73.65}} \\
  \hline
   \end{tabular}
  \end{adjustbox}
  \caption{Comparison of IoU (\%) across classes for various methods on custom data splits SPLIT4 of the ISPRS Potsdam dataset.}
  \label{table:pots_result}
 \end{minipage}
\end{table*}

\section{Result and Ablations}

\subsection{Baselines}
Unfortunately, we were unable to compare the performance of the proposed method with any UDA method based on geographical considerations because the code for the previous SOTA method for geographic SS \cite{10208468} is unavailable. Nevertheless, we compared our approach to the results reported by four state-of-the-art UDA methods for land cover mapping: AdaptSegNet \cite{Tsai_adaptseg_2018}, CLAN \cite{clan}, IAST \cite{iast}, and ADVENT \cite{vu2019advent}. Additionally, we evaluated our results for SPLIT1 (\cref{table:splits}) against SOTA GeoMultiTaskNet \cite{10208468} on the FLAIR \#1 dataset.

\subsection{Experimental Setup}
For our implementation, we applied the settings common to most works \cite{ding2019semantic,zhang2018context,zhu2019asymmetric,li2019expectation} with batch size 16, 0.001 initial LR, and poly learning decay with Adam during training. In addition, during training, the images are randomly cropped from $512 \times 512$ to $256 \times 256$, while inference is performed over all patches that are merged back to their original shape. 
The training iteration was set to 200 iterations unless otherwise specified. We used early stopping with a patience of 30 epochs. Additionally, we set the constant value of random seed as 18 for different modules/packages used during implementation. A single NVIDIA GeForce RTX 4090 GPU (24 GB) was used in the training phase. 
Similar to the works discussed above, we chose the standard mean intersection over union (mIoU) as our evaluation metric.

In \cref{table:comparison},
we present a comparison of class-wise Intersection over Union (IoU) between the baseline models and our model. All the metrics reported in \cref{table:comparison} and  \cref{table:pots_result} are either obtained directly from the previous published sources or are obtained by training their official implementation along using their pretrained weight.
To evaluate the performance of our model across different domain gaps, we extended our experiment by creating 2 additional custom data splits of FLAIR\#1 dataset and 1 from ISPRS Potsdam dataset ( mentioned in \cref{table:splits}) and obtained the results listed in ~\cref{table:table2}--\cref{table:table3}
and \cref{table:pots_result}. Experiments were conducted to investigate the behavior of SegDesicNet across different values ($\lambda_{\text{min}}$, $\lambda_{\text{max}}$) and numbers of grid scales ($S$) alongside the domain contribution factor $\alpha$. The results for SPLIT1 are presented in \cref{table:hyper}. We used this specific value for all of our experiments. 
Overall, the results on the target domain test set of FLAIR \#1 and ISPRS Potsdam demonstrate that the identified lightweight networks possess the UDA-oriented capability to transfer knowledge from a source domain to a target domain.
Our results demonstrate a considerable improvement in the mean IoU (mIoU) by approximately 6\% compared with state-of-the-art performance (refer to \cref{table:table1}). Similar findings were obtained for the custom domain split of the ISPRS Potsdam dataset. Our proposed algorithm outperformed the other baseline methods. It is also worth noting that ranking of the different methods varies between the various domain splits, and some models failed to recognize certain classes entirely, our methods consistently outperformed them and maintained a very consistent final value. Notably, unlike the approach described in \cite{10208468}, we did not employ any specific weighting mechanism or any data augmentation and designed a very lightweight method with approximately 27\% fewer model parameters, despite this, our proposed method achieved robust performance across nearly all underrepresented classes (including \textit{coniferous, deciduous, brushwood, vine, grassland, crop, and plowed land}) among every data splits.

\begin{table}[hbt!]

 \small
 \renewcommand{\arraystretch}{0.9}
 \begin{tabular}{>{\centering\arraybackslash}m{1.7cm}>{\centering\arraybackslash}m{1.0cm}>{\centering\arraybackslash}m{1.0cm}>{\centering\arraybackslash}m{1.2cm}>
 {\centering\arraybackslash}m{1.2cm}}
  \toprule
  \textbf{$\lambda_{\text{min}}$} & \textbf{$\lambda_{\text{max}}$} & \textbf{$S$} & \textbf{$\alpha$} & \textbf{$mIoU$} \\
  \midrule
  --    & --     & -- & 0& 45.6 \\
  0.01   & 0.00001  & 8  &0.5 & 47.9 \\
  0.01   & 0.000001  & 8  & 1.0 & 45.52 \\
  \textbf{0.01 }  & \textbf{0.00001 }  & \textbf{16} & \textbf{0.5} & \textbf{50.12} \\
  0.1   & 0.00001  & 16 & 0.5 & 49.46 \\
  0.01   & 0.000001  & 16 &1 & 45.52 \\
  0.01   & 0.00001  & 32 &0.5 & 48.75 \\
  0.1   & 0.00001  & 32 & 0.5& 49.07 \\
  0.01   & 0.000001  & 32 & 1 & 45.48 \\
  0.01   & 0.00001  & 64 & 0.5& 48.60 \\
  0.01   & 0.000001  & 64 &1 & 46.66 \\
  0.1   & 0.0001   & 64 &1 & 48.6 \\
  \bottomrule
 \end{tabular}
 \caption{Observations depicting the performance of SegDesicNet across various hyperparameters for SPLIT1}
 \label{table:hyper}
\end{table}
\begin{figure*}[htb]
  \centering
  \includegraphics[width=1.05\textwidth]{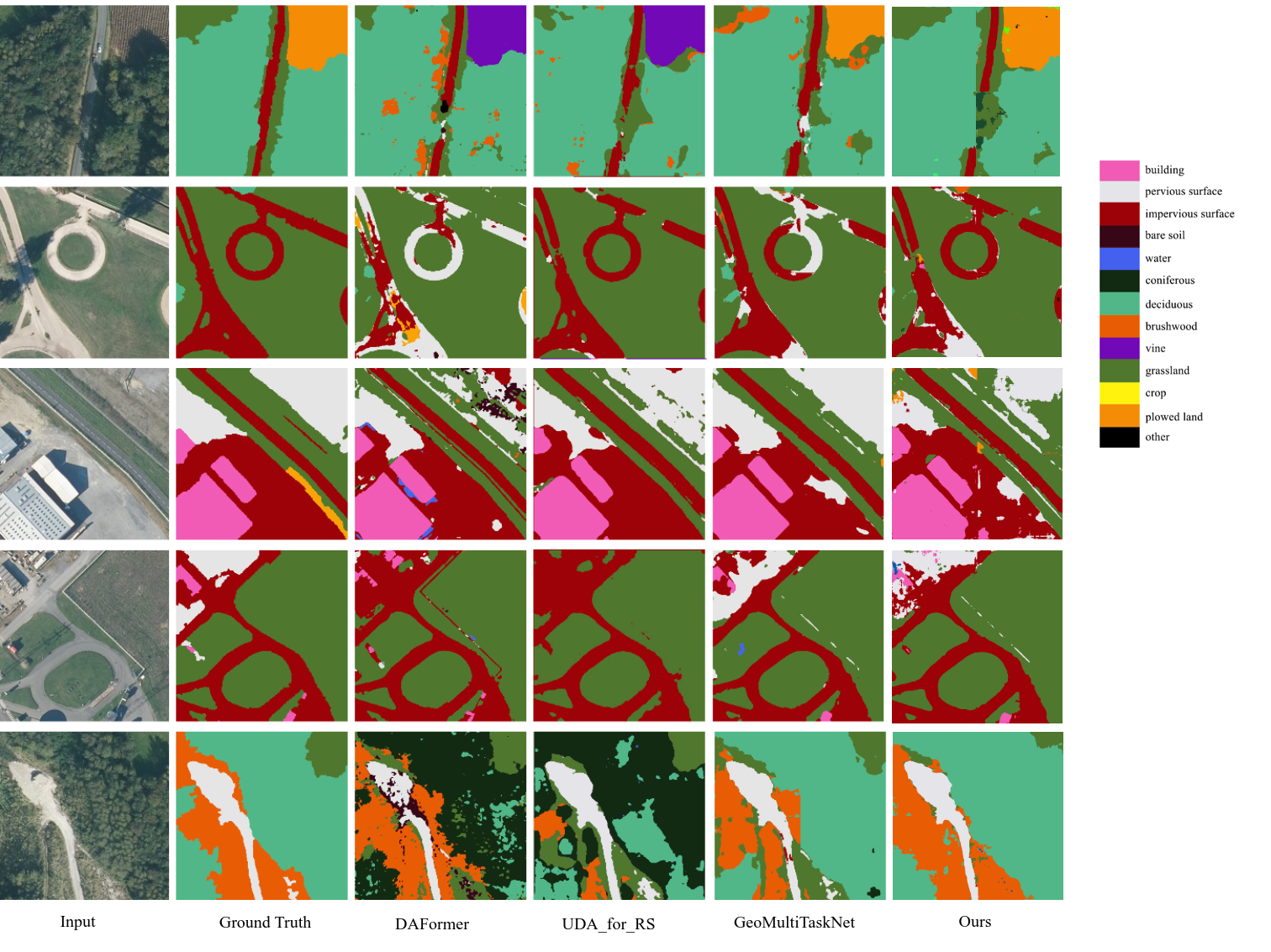}
  \caption{Comparison of segmentation map prediction on the target domain images of SPLIT1 by baseline models to our best model. }
  \label{fig:example}
\end{figure*}
\textbf{Qualitative Evaluation :}In \cref{fig:example}, examples of land cover mapping results on the target domain of FLAIR \#1 SPLIT1. In this figure, we observe that while GeoMultiTaskNet comes close to our results for classes such as \textit{plowed land} and \textit{brushwood}, our proposed method outperforms it considerably. Our model successfully captured distinctive features that might vary in style (visual features), as seen in the second image. This indicates our model's ability to handle style differences in mobility infrastructure, which can occur owing to large domain differences, a key challenge in domain adaptation.
However, our model has shown some indistinctness in differentiating \textit{pervious surfaces} from \textit{grasslands}. Despite this minor shortcoming, it still performs exceptionally well across other categories. Notably, DAFormer and UDA\_for\_RS, despite their strong performance in the impervious surface class, exhibited misclassification issues. This highlights the robustness of our model in maintaining accuracy across a broader range of land cover types.


\section{Limitation and Future Works}

Although the proposed method demonstrated encouraging performance, it has some limitations. First, there are several classes, particularly the \textit{pervious surface} class, which consistently underperformed throughout the experiment. As shown in \cref{fig:example}, the model performed well when the \textit{pervious surface} class was surrounded by a \textit{deciduous} class. However, its performance degrades when \textit{pervious surface} class is surrounded by \textit{grassland}. All the baselines including our methods has shown low accuraccy over the pixels affected by shadows. This observation suggests that there is room to improve the proposed model by incorporating a class-conditional style in which the probability of a class can be further analyzed based on its neighboring classes.
In addition, during training, we chopped the input images to a size of $256\times 256$ in an attempt to decrease the network parameter, but we can see in the \cref{fig:example} top right that our impervious surface is misjudged due to non overlapping splitting. Hence, better conditional cropping should be performed to address it in better terms. Second, to narrow the domain gap between the neural network's style of understanding and the artificial neural network we used, we implemented GRID positional encoding. However, this approach may not be optimal. There is a pressing need to explore more efficient methods for modeling geocoordinates that account for the Earth's curved surface to make our approach more human-centric. In addition, one constraint of lightweight modeling in the context of remote sensing data is the necessity for real-time execution across various applications. we addressed this by manipulating our network layers, there may be numerous possibilities to address this more effectively. Also more features can be added to this framework,  which includes enabling the model to utilize on cross-sensor data, such as by predicting the height of a pixel along with its classes. Thus, we can enhance the practicality and applicability of the proposed method in real-world scenarios.

\section{Conclusion}

Our work has explored unsupervised domain adaptation methods using the geographical coordinates of image patches. Our findings demonstrate that incorporating geo-coordinates into neural network training enhances the model's ability to adapt to target domains without ground truth label, leading to improved performance. This innovative method has pivotal implications for various diverse environmental applications, as it enables accurate and efficient adaptation due to incorporation geo-coordinates. By mimicking human perception of the physical environment, our approach advances the understanding of how neural networks perceive their surroundings. When considering latitude and longitude values on a spherical surface, cosine dissimilarity is a more effective approach compared to Euclidean distance or mean squared error (MSE) loss. This is because cosine dissimilarity takes into account the Earth's curved surface, unlike other distance metrics that assume a flat Euclidean space, cosine dissimilarity captures the spherical nature of the Earth, resulting in a more comprehensive evaluation. Although our research showed promising results, it is crucial to acknowledge some limitations. Future research could investigate alternative methods for incorporating geographical information into neural network architectures for unsupervised domain adaptation. Furthermore, assessing the scalability and computational efficiency of the proposed method across various domains and datasets can enhance its practical application. Our study demonstrates the potential of utilizing geographical coordinates to enhance unsupervised domain adaptation in neural networks. By learning from the human perception of their physical surroundings, our approach presents a promising avenue for addressing the challenges in adapting models to new environments. By embracing interdisciplinary approaches and incorporating insights from fields such as geography and cognitive science, we can continue to advance machine learning and deepen our understanding of how neural networks interact with the physical world.

\section*{Acknowledgements}
\vspace{-0.5em}
\noindent This research received funding from the PERSEUS project, a European Union’s Horizon 2020 research and innovation program under the Marie Skłodowska-Curie grant agreement No 101034240. This paper is supported by the MoST (MobilitetsLab Stor-Trondheim) project (\url{https://www.mobilitetslabstortrondheim.no/en/})
\newpage
{\small
\bibliographystyle{ieee_fullname}
\bibliography{main}
}

\end{document}